\documentclass{article}

\usepackage{style}
\usepackage[utf8]{inputenc}
\usepackage[T1]{fontenc}
\usepackage{microtype}
\usepackage{newtxtext}
\usepackage{fancyvrb}
\usepackage{enumitem}

\usepackage{hyperref}
\usepackage{url}
\usepackage{doi}

\hypersetup{
    colorlinks=true,
    linkcolor=blue,
    filecolor=blue,
    urlcolor=blue!70!black,
    citecolor=blue,
    pdfborder={0 0 0}
}

\usepackage{booktabs}
\usepackage{amsfonts}
\usepackage{nicefrac}
\usepackage{amsmath}
\usepackage{tabularx}
\definecolor{green}{RGB}{102, 217, 129}
\definecolor{blue}{RGB}{110,193,228}
\definecolor{yellow}{RGB}{255,245,102}
\definecolor{orange}{RGB}{255,169,64}
\definecolor{red}{RGB}{255,77,79}
\colorlet{severitylow}{blue!70!white}
\colorlet{severitymoderate}{yellow!70!white}
\colorlet{severityhigh}{orange!70!white}
\colorlet{severitycritical}{red!70!white}
\colorlet{normalbehavior}{green!70!white}

\usepackage{graphicx}
\usepackage{subcaption}

\usepackage{natbib}
\setcitestyle{square,numbers,sort&compress}

\newcommand{\formatkeywords}{%
  \begingroup
  \def\tempdo{0}%
  \def\do##1{\if\tempdo0\else\ \and\fi##1\def\tempdo{1}}%
  \expandafter\docsvlist\expandafter{\keywordslist}%
  \endgroup
}

\newcommand{\firstpagefooter}[1]{%
    \fancypagestyle{footer}{%
        \fancyhf{}%
        \fancyfoot[C]{%
            \small\color{black!60}%
            \begin{minipage}[t]{\textwidth}%
                \vspace{-1.5cm}%
                \raggedright%
                \textcolor{black!30}{\rule{2cm}{0.3pt}}\\[0.2em]%
                #1%
            \end{minipage}%
        }%
    }%
}
\firstpagefooter{First draft: \firstdraft; Last revision: \lastrevision.}

\newcommand{\doctitle}{A Feature Engineering Approach for Business Impact-Oriented Failure Detection in Distributed Instant Payment Systems}
\newcommand{\docauthor}{Lorenzo Porcelli}
\newcommand{\docsubject}{cs.DC, cs.LG, cs.AI, eess.SP}
\newcommand{\keywordslist}{AIOps, Observability, Failure Management, Explainable Anomaly Detection, Distributed Systems}

\newcommand{\firstdraft}{23 November 2023}
\newcommand{\lastrevision}{2 August 2025}

\title{\doctitle}

\author[a,b]{\orcidicon{0009-0004-2539-2431}Lorenzo Porcelli}

\affil[a]{Directorate General for Information Technology, Bank of Italy, Frascati, Italy}
\affil[b]{Department of Computer Science, University of Salerno, Fisciano, Italy}

\hypersetup{
  pdftitle={\doctitle},
  pdfauthor={\docauthor},
  pdfsubject={\docsubject},
  pdfkeywords={\keywordslist},
  colorlinks=true,
  linkcolor=blue,
  citecolor=blue
}

\begin{document}
\maketitle
\thispagestyle{footer}

\begin{abstract}
Instant payment infrastructures have stringent performance requirements, processing millions of transactions daily with zero-downtime expectations. Traditional monitoring approaches fail to bridge the gap between technical infrastructure metrics and business process visibility. We introduce a novel feature engineering approach based on processing times computed between consecutive ISO 20022 message exchanges, creating a compact representation of system state. By applying anomaly detection to these features, we enable early failure detection and localization, allowing incident classification. Experimental evaluation on the TARGET Instant Payment Settlement (TIPS) system, using both real-world incidents and controlled simulations, demonstrates the approach's effectiveness in detecting diverse anomaly patterns and provides inherently interpretable explanations that enable operators to understand the business impact. By mapping features to distinct processing phases, the resulting framework differentiates between internal and external payment system issues, significantly reduces investigation time, and bridges observability gaps in distributed systems where transaction state is fragmented across multiple entities.
\end{abstract}
\keywords{\formatkeywords}

\section{Introduction}  \label{sec:introduction}
The SEPA Instant Credit Transfer (SCT Inst) scheme, launched by the European Payments Council in 2017, represents a significant harmonization effort for instant payment systems across Europe. The SCT Inst scheme provides a maximum execution time of 10 seconds from payment initiation by the originator bank to settlement notification receipt, with around-the-clock availability and immediate fund accessibility to the beneficiary~\cite{epc2022sctinstrulebook}. Currently, two infrastructures serve as instant payment settlement systems in Europe: RT1~\cite{rt1} and TARGET Instant Payment Settlement (TIPS)~\cite{renzetti2021tips}, which process millions of transactions daily, with volumes expected to surge following the implementation of the Instant Payment Regulation (IPR)~\cite{targetannualreport2023,ipr}.

The technical requirement for TIPS is to process the settlement of each individual payment within 5 seconds. To meet this performance requirement with zero-downtime expectations, TIPS is designed as a distributed system~\cite{caricato2022tips}. Its decentralized nature, while offering benefits in scalability and resource utilization, introduces significant operational complexity through increased abstraction layers and component interdependencies. Issues in instant payment systems directly affect financial institutions, businesses, and consumers through immediate financial and reputational impacts.

Systems such as TIPS require monitoring approaches that can detect failures---undesirable deviations in system functioning, such as processing delays---before they escalate into incidents. Traditional monitoring tools face fundamental limitations, as they rely on manual investigation processes and fail to establish meaningful connections between infrastructure metrics and business processes. Operating in isolated data silos, these tools cannot contextualize technical metrics within the broader settlement process, making real-time business impact assessment challenging.

Information Technology (IT) operations teams are overwhelmed by the volume of monitoring data requiring real-time analysis, creating unmanageable workloads even for specialized operators. Operations and maintenance solutions that rely heavily on human monitoring and manual intervention are inadequate for rapidly and efficiently managing emerging issues in real-time systems.

Recent research~\citep{dang2019aiops,nedelkoski2019anomaly} has focused on developing intelligent software systems to address IT operations challenges by inferring a system's internal state from its external outputs---a concept known as observability. These approaches, grouped under the term Artificial Intelligence for IT Operations (AIOps), leverage data-driven technologies such as Machine Learning, Big Data, and Data Mining to enhance IT operations~\citep{notaro2021survey}. However, current AIOps solutions are not a panacea, as their effectiveness depends on training data quality, and they require trust in tools whose outputs may lack explainability. Additionally, they introduce further infrastructure complexity and incur high implementation and maintenance costs.

In the context of instant payments, processing times across distributed system components offer a unique observability opportunity. By considering message volumes and computing processing durations between consecutive messages that constitute a transaction, we build features representing the holistic operational state of the payment infrastructure. We treat failure detection as an anomaly detection problem~\cite{chandola2009anomaly}, assuming that failures generate irregular behaviors in payment systems by increasing processing times or reducing settled payment volumes. We derive a normal behavior model from multivariate time series and test new observations against this baseline to identify deviations.

Our research contributions include:
\begin{itemize}
    \item a novel feature engineering approach that computes processing times between consecutive ISO 20022 message exchanges to create a compact representation of payment system state;
    \item a formal definition of the failure detection problem in instant payment infrastructure from the perspective of a Clearing and Settlement Mechanism (CSM);
    \item an anomaly detection framework that employs computed processing times to bridge the gap between infrastructure observability and business process visibility, treating detected anomalies as indicators of potential system failures.
\end{itemize}

By analyzing detected anomalies, the proposed approach determines whether failures have caused incidents and estimates business impact by measuring effects on operations and processes. This capability is achieved by integrating domain knowledge that enables failure localization and incident classification. The framework complements traditional monitoring systems by providing early detection of service degradation invisible at the individual component level, while delivering explanations that directly guide remediation efforts before issues escalate into severe incidents.

The remainder of this paper is structured as follows. Section~\ref{sec:related_work} reviews the state of the art in AIOps, explainable anomaly detection, and feature engineering techniques for distributed systems, highlighting research gaps in instant payment monitoring. Section~\ref{sec:sct_inst} provides essential background on the SEPA Instant Credit Transfer scheme and its standardized message exchange patterns. Section~\ref{sec:methodology} presents our feature engineering approach for failure detection in instant payment systems. Section~\ref{sec:experiments} evaluates this approach on both real-world incidents and controlled simulations on the TARGET Instant Payment Settlement system. Finally, Section~\ref{sec:conclusion} concludes the paper.

\section{Related Work} \label{sec:related_work}
Machine learning applications in financial systems have predominantly focused on fraud detection, with research attention primarily directed toward business aspects such as credit and debit card fraud, insurance fraud, and money laundering, rather than operational failure detection in system infrastructures~\citep{hilal2022financial, diadiushkin2019fraud}. Our research positions itself at the intersection of Artificial Intelligence for IT Operations (AIOps), explainable anomaly detection techniques, and feature engineering approaches in distributed systems.

\subsection{Artificial Intelligence for IT Operations}
Artificial Intelligence for IT Operations (AIOps) lacks a universally accepted definition, remaining a largely unstructured, cross-disciplinary area of study. Despite definitional variations, AIOps commonly aims to provide complete visibility into IT system operational states, thereby enhancing customer services~\cite{notaro2021survey}. Research in this emerging field typically focuses on two primary domains: failure management~\cite{kobbacy2007ai,mukwevho2021toward}, which encompasses techniques to minimize failure occurrence and impact in large-scale systems, and resource provisioning~\cite{dang2019aiops,nedelkoski2019anomaly}, which optimizes the allocation of power, compute time, network bandwidth, and virtual memory resources.

Failure management approaches are conventionally categorized as proactive (failure avoidance) or reactive (failure tolerance)~\cite{notaro2021survey}. Failure avoidance addresses potential failures before their occurrence through predictive analysis of system state or preventive actions to reduce future incidence~\cite{li2017software,xiao2018disk}. Conversely, failure tolerance manages errors after manifestation to assist operators and improve mean time to recovery, encompassing failure detection~\cite{xu2018unsupervised,audibert2020usad,su2019robust}, root cause analysis (RCA)~\cite{li2019generic,lin2020fast}, and remediation processes~\cite{wang2017constructing,lin2018hardware}.

Complex microservice architectures particularly benefit from AIOps approaches that leverage multimodal data sources---including logs, metrics, and traces---for more effective fault localization and diagnostics~\cite{zhang2024failure}. Multivariate log-based frameworks can aggregate features from distributed nodes to enable cluster-wide anomaly detection~\cite{zhang2025logdb}, while integrated approaches combining these data sources provide more precise diagnostic capabilities~\cite{wang2025tamo}. Beyond diagnostics, AI-driven predictive failure analysis has evolved from statistical methods to sophisticated deep learning architectures, including Long Short-Term Memory (LSTM), Transformer, and Graph Neural Networks (GNN), enabling proactive maintenance and automated recovery strategies~\cite{kercheval2015modelling,polu2024ai}.

A promising research direction has emerged in leveraging Large Language Models (LLMs) to enhance AIOps capabilities, particularly for fault diagnosis and RCA. LLM-powered on-call systems can automatically match incidents to handlers, aggregate diagnostic information, predict root causes, and provide explanatory narratives~\cite{chen2024automatic}. Specialized language models trained on IT-related data and employing mixture-of-adapter strategies offer efficient tuning across different operational domains~\cite{guo2023owl}. Holistic frameworks for incident management with network-specialized LLMs effectively decompose complex operational tasks~\cite{hamadanian2023holistic}. Systematic evaluation of these approaches requires comprehensive frameworks for end-to-end assessment of AIOps agents through controlled fault injection and agent-cloud interactions~\cite{shetty2024building,chen2025aiopslab}.

\subsection{Explainable Anomaly Detection}
Explainable AI (XAI) encompasses methods and techniques that enable human users to understand and trust the outputs created by machine learning algorithms~\cite{doshi2017towards}. Although the literature lacks a clear distinction between interpretability and explainability, systems that provide understanding to designers are typically considered interpretable, while those that make outputs comprehensible to users are deemed explainable~\cite{miller2017explainable}. When applied to anomaly detection, these techniques constitute eXplainable Anomaly Detection (XAD).

Financial infrastructures require explainable AI systems not merely as a desirable feature but as a fundamental operational necessity. This requirement stems from multiple imperatives: building user trust, enabling effective debugging of both AI models and monitored systems, identifying potential biases, and meeting increasingly stringent regulatory compliance mandates. From a system management perspective, clear explanations enable operators to assess alert criticality, reduce false positive fatigue, and take appropriate corrective actions. The acknowledged struggle with explainability in existing AIOps solutions highlights the need for continued innovation in this domain.

XAD techniques can be categorized based on their position in the detection pipeline as pre-model, in-model, or post-model approaches~\cite{li2023survey}. Pre-model techniques operate before anomaly detection begins, including feature selection~\cite{noto2012frac,pang2016outlier,pang2016unsupervised} and feature representation methods~\cite{ramirez2019computational,schlegl2021scalable,dissanayake2021robust}. These approaches can simplify models and enhance accuracy, but may not fully capture complex data relationships.
In-model techniques use inherently interpretable models for anomaly detection. These include transparent models in supervised learning~\cite{lipton2018mythos}, feature subset-based models~\cite{savkli2021random}, and other models in unsupervised learning that provide clarity by design~\cite{smets2011odd}. 

Post-model techniques examine anomaly detection models after processing completion and divide into shallow and deep approaches. Shallow post-model techniques include subspace-based methods~\cite{angiulli2009detecting,macha2018explaining} and surrogate models~\cite{lundberg2017unified,ribeiro2016should}. Deep post-model techniques are aimed at interpreting anomalies detected by deep learning models, such as Autoencoders, Recurrent Neural Networks, Convolutional Neural Networks, and other Deep Neural Networks~\cite{chong2021toward,herskind2021outlier,xu2021beyond}. LLMs can further enhance explainability. For example, they can be used to generate narratives alongside root cause predictions, leading to human-understandable explanations in operational contexts~\cite{chen2024automatic}.

\subsection{Feature Engineering Techniques in Distributed Systems}
The effectiveness of anomaly detection in distributed systems depends critically on feature engineering---the process of extracting informative representations from raw data. This process presents unique challenges in distributed environments where data is generated across numerous nodes, services, or databases, often at high velocity and volume. The complexity becomes particularly evident in transactional data, where capturing evolving states, dependencies, and temporal sequences remains crucial while maintaining consistency across distributed sources.

Telemetry sources in distributed systems include logs, metrics, and traces, each offering different insights into system behavior. Log parsing, a preliminary step in feature extraction, has been significantly enhanced through LLMs. The in-context learning capabilities of models like GPT-3 now enable the generation of log templates without training. These models use diversity-based candidate sampling to achieve state-of-the-art parsing accuracy across multiple public datasets~\cite{xu2024divlog}. Few-shot tuning approaches with various LLMs effectively conceptualize log parsing as a ``translation'' task from raw logs to templates~\cite{ma2024llmparser}. Building on these parsing techniques, log events can be enriched using LLMs with fine-tuned BERT models to produce event embeddings that serve as features for downstream transformer-based detection models~\cite{he2024llmelog}.

Modern distributed systems demand that features be generated, processed, and made available to detection models almost instantaneously, often during transaction processing or immediately after. This requirement eliminates many traditional batch-oriented feature engineering techniques and necessitates stream processing capabilities. LogDB utilizes LSTM networks with self-attention to compress and aggregate sequence patterns, event counts, and semantic information from logs at master nodes, resulting in informative node-level representations for distributed log data feature engineering~\cite{zhang2025logdb}. Metric data from individual components provides another valuable source for feature extraction in distributed systems in real time, with techniques that detect distinctive patterns to identify problematic machines~\cite{deng2025minder}. Similarly, features derived from communication operators effectively identify performance irregularities in complex processing environments~\cite{yao2025holmes}.

\subsection{Research Gaps and Contributions}
Existing feature extraction approaches for telemetry sources predominantly focus on system component-level behavior. However, financial transactions such as instant payments traverse multiple organizational boundaries throughout their lifecycle. Limiting observability to data generated within organizational perimeters creates fundamental gaps, where each organization observes only fragments of the complete process, thereby impeding effective anomaly localization. Current monitoring systems exemplify this limitation by concentrating on component logs and metrics while failing to identify issues originating beyond infrastructure boundaries.

We address these gaps by introducing domain-specific features extracted from transaction message flows, enabling operators to correlate technical anomalies with business impact. Variations in transaction-derived time series serve as early indicators of potential system degradation across the entire SEPA Instant Credit Transfer (SCT Inst) process. Our features are tied to standardized SCT Inst process traces rather than product-specific telemetry and are designed to map distinct processing phases. This design enables early failure detection, failure localization in distributed systems, and incident severity classification for business impact estimation. The approach maintains robustness against software updates and vendor changes that may render telemetry data obsolete.

While AIOps anomaly detection effectiveness depends critically on training data quality, output comprehensibility relies equally on feature significance. Real-time computational constraints render post-hoc XAI techniques based on LIME~\cite{ribeiro2016should} or SHAP~\cite{lundberg2017unified} impractical for high-dimensional data. This creates scenarios where technical anomalies are detected but their significance, and therefore business impact, remains opaque to operators. Our framework prioritizes explainability by design through the proposed feature engineering approach, providing causal rather than correlational explanations. The subsequent section introduces SEPA Instant Credit Transfer schema fundamentals necessary for understanding our feature engineering approach.

\section{The SEPA Instant Credit Transfer Scheme} \label{sec:sct_inst}
The proposed feature engineering approach derives features from traces of messages exchanged during the SEPA Instant Credit Transfer (SCT Inst)~\citep{epc2022sctinstrulebook}. This scheme establishes rules, practices, and standards that enable fund transfers to beneficiary accounts. The settlement process involves at least five actors. At the endpoints are the Originator (Payer) and the Beneficiary (Payee), while between them operate at least three intermediaries: the Originator Payment Service Provider (PSP), the Beneficiary PSP, and the Clearing and Settlement Mechanism (CSM).

The Originator initiates the transaction by instructing an SCT Inst from the Originator PSP. The Beneficiary, holding an account at the Beneficiary PSP, is the ultimate recipient of the transferred funds. The CSM, which may comprise one or more entities, collectively performs the clearing and settlement functions. Clearing encompasses the stages of information exchange and determination of a final settlement position, while settlement involves the extinction of obligations defined during the clearing stage. A credit transfer can only be executed if the PSPs of the Originator and Beneficiary adhere to the same CSM or to different but interoperable CSMs.

In a simple scenario where two PSPs join the same CSM, the SCT Inst settlement process starts when the Originator requests the Originator PSP (typically a commercial bank) to execute an SCT Inst. After reserving the transfer amount in the Originator's account, the Originator PSP transmits the order to the CSM. The latter dispatches the instruction to the Beneficiary PSP, awaiting acceptance or rejection of the payment. The Beneficiary PSP performs a series of checks before sending an acceptance or rejection notification to the CSM, which then relays it back to the Originator PSP. The subsequent steps depend on whether the payment is accepted. In the case of acceptance, the Beneficiary PSP makes the received funds available in the Beneficiary's account, and the Originator PSP debits the reserved amount from the Originator's account. Conversely, if the payment is rejected, the Originator PSP must inform the Originator of the Beneficiary PSP's decision.

The current implementation of SCT Inst is based on ISO 20022, an open standard for defining financial data content~\citep{epc2022sctinstrulebook}. ISO 20022 has gained global prominence, with many of the world's largest Securities Financial Market Infrastructures already adopting it. Since the end of 2022, all institutions sending or receiving payment messages via SWIFT have been transitioning to ISO 20022, a trend followed by numerous Payments Market Infrastructures worldwide~\citep{iso20022fordummies}.

\section{A Feature Engineering Approach for Failure Detection} \label{sec:methodology}
We treat failure detection as an anomaly detection problem, where engineered features capture the operational state of the system and anomaly identification serves as an early warning mechanism for preventive interventions before significant service degradation occurs. This section formalizes the anomaly detection problem from a Clearing and Settlement Mechanism (CSM) perspective, introduces our feature engineering approach that derives a compact system state representation from processing times between consecutive ISO 20022 messages, and demonstrates how these features enable explainable detection with failure localization and incident severity classification capabilities that directly inform business impact assessment.

\subsection{Problem Formulation} \label{sec:ad_problem}
From the perspective of a Clearing and Settlement Mechanism (CSM), an anomaly in instant payment processing represents any significant deviation from normal operational behavior. The most critical anomaly is typically a substantial drop in expected payment volume, which indicates the platform's diminished ability to fulfill its core service function.
More generally, the $i$-th observation of a time series of settled instant payments $S^{\eta}$, observed at frequency $\eta$, is considered anomalous if at least one of its components deviates significantly from the expected value at time $\tau_i$.

The fundamental challenge addressed in this research is to determine, at time $\tau_i + \varepsilon$, where $\varepsilon$ is a negligible time interval, whether the current operational state of the instant payment system is anomalous within the SEPA Instant Credit Transfer (SCT Inst) process context, and if so, to identify which infrastructure component is the origin of this anomaly and assess its severity.

We address this challenge by decomposing it into two distinct subproblems:
\begin{enumerate}
    \item estimating the unknown function $f$ that determines whether an observation of time series $S^{\eta}$ belongs to the set of anomalies $A$;
    \item given a detected anomaly $a \in A$, estimating the unknown function $g$ that determines whether $a$ belongs to the set of anomalies originating within the CSM infrastructure $A_1 \subseteq A$ or to the set of anomalies caused by external factors $A_2 \subseteq A$, and classifying the severity of any incident that may result from $a$.
\end{enumerate}

Upon completing these two phases, we obtain explanations for identified anomalies that guide IT operators' investigations by providing insights into the distributed system's state. For instance, set $A_1$ encompasses misconfigurations or hardware failures within the CSM infrastructure, while set $A_2$ includes external factors such as network issues or participant-related problems. The combination of detected anomalies and their corresponding scores enables incident severity classification and business impact assessment.

The following sections formalize how we represent system state through time series observations of settled payments.

\subsection{Real-time Feature Engineering for Instant Payments}
To capture relevant operational patterns of the instant payment system that enable preemptive detection of failures, we need a system state representation that can be computed efficiently. To this end, we introduce a novel feature engineering approach based on processing times computed between consecutive ISO 20022 message exchanges.

\subsubsection{Processing Times as Features}
From a technical perspective, an instant payment transaction consists of standardized messages exchanged between participating entities. The CSM is involved twice during processing: first in the conditional phase (validation, fund reservation, and notification to the beneficiary participant) and later in the settlement phase. The relevant ISO 20022 messages are pacs.008 and pacs.002, where pacs.008 initiates a request while pacs.002 responds to it.

\begin{figure*}
    \centering
    \includegraphics[width=.95\textwidth]{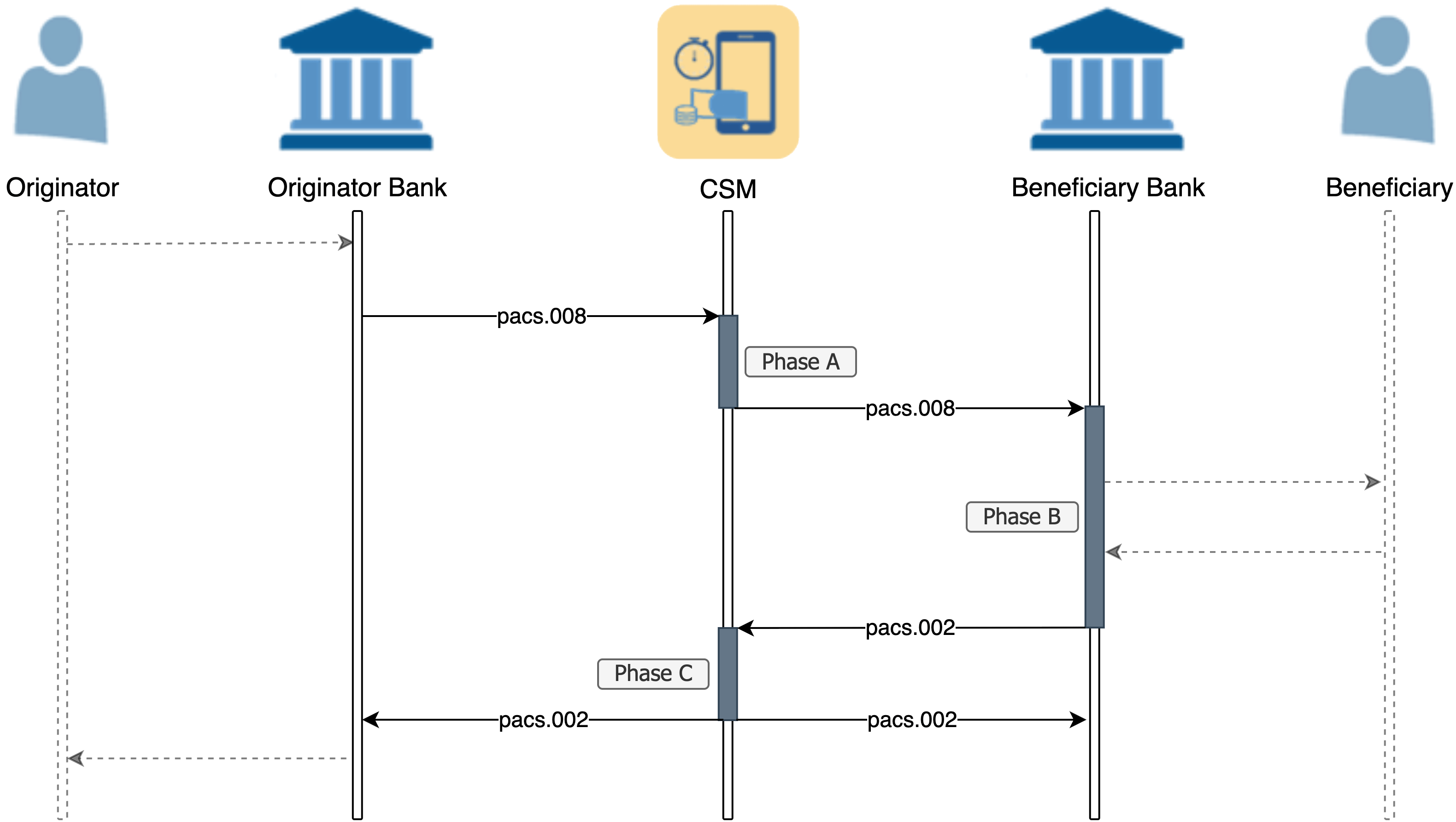}
    \caption{\small The core message flow of the SCT Inst process showing the three processing phases identified in the SEPA Instant Credit Transfer process, whose durations constitute the proposed features for anomaly detection.}
    \label{fig:sct_inst_process}
\end{figure*}

In the standard settlement model, a transaction can have four possible final states: settled, expired due to timeout, rejected by the beneficiary, or failed due to validation errors or insufficient liquidity. Our conceptual model considers only settled payments, as these provide sufficient information to establish a baseline model and allow calculation of all three processing time features. As illustrated in Figure~\ref{fig:sct_inst_process}, we define three processing phases: the duration of the conditional phase within the CSM (Phase A), the cumulative time spent outside the CSM awaiting the beneficiary's reply (Phase B), and the time within the CSM for the settlement phase (Phase C).

We formally define an instant payment as a tuple 
\begin{equation}
    \mathbf{p} = (\delta_1, \delta_2, \delta_3),
\end{equation}
where:
\begin{itemize}
    \item $\delta_1$ approximates Phase A duration, computed as the time difference between the CSM sending the pacs.008 message to the beneficiary bank and the CSM receiving the pacs.008 message from the originator bank;
    \item $\delta_2$ approximates Phase B duration, computed as the time difference between the CSM receiving the pacs.002 message from the beneficiary bank and the CSM sending the pacs.008 message to the beneficiary bank;
    \item $\delta_3$ approximates Phase C duration, computed as the time difference between the CSM sending the pacs.002 messages to both the originator and beneficiary banks and the CSM receiving the pacs.002 message from the beneficiary.
\end{itemize}

These processing times provide valuable operational insights into the system's behavior. The sum $\sum_{i=1}^{3} \delta_{i}$ estimates a lower bound for the total transaction processing time, while $\sum_{i\in\{1,3\}} \delta_{i}$ represents the time spent across CSM components during the entire process. For instance, unexpected increases in processing times $\delta_1$ or $\delta_3$ may indicate failures in internal CSM infrastructure components, while anomalies in $\delta_2$ processing times could suggest external failures involving participants or network connectivity problems.

\subsubsection{Data Aggregation and Time Series Representation}
Since instant payments occur at irregular intervals, we transform individual transaction data into a regular time series that includes both processing time information and settled payment volume information.

We define a time series of $n$ settled transactions as:
\begin{equation}
    S = \{ (t_i, \mathbf{p}_i) \}_{i=1}^{n} = \{ (t_i, {\delta_1}_i, {\delta_2}_i, {\delta_3}_i) \}_{i=1}^{n},
\end{equation}
where the $i$-th observation $(t_i, \mathbf{p}_i)$ represents the instant payment $\mathbf{p}_i$ settled at time $t_i$.

To create a regular time series representation, we resample this irregular series into $m$ bins---that is, we aggregate the data into fixed time intervals---using a fixed sampling rate $\eta$, where the number of bins is determined by $m = \lceil (\omega - \alpha)/\eta \rceil$ with $\alpha \leq t_1$ as the start timestamp and $\omega \geq t_n$ as the end timestamp. The resampled time series is defined as:
\begin{equation} \label{eq:ts_settled_ip_resampled}
    S^{\eta} = \big\{ (\tau_k, \tilde{\mathbf{p}}_k, v_k ) \big\}_{k=1}^{m}
\end{equation}

Within this resampled series, each bin $b_k$ covers the time interval $\big[ \alpha + (k-1) \eta, \ \alpha + k \eta \big)$, and the corresponding aggregated observation is defined by:
\begin{itemize}
    \item $\tau_k = \alpha + (k-1) \eta$ is the representative timestamp for the $k$-th bin;
    \item $\tilde{\mathbf{p}}_k$ is the aggregated value of the processing times for instant payments within bin $b_k$, with the aggregation function being either the mean or the median;
    \item $v_k = \lvert b_k \rvert$ represents the number of payments within bin $b_k$.
\end{itemize}

\subsection{Explainable Anomaly Detection Framework} \label{sec:xad}
The anomaly detection framework consists of two complementary components: a detector that identifies potential system failures by assigning anomaly scores to aggregated observations, and an explainer that characterizes these detected anomalies by determining their severity and localization, enabling business impact assessment.

\subsubsection{The Anomaly Detector}
To streamline notation for the anomaly detection process, we redefine each aggregated observation of the time series $S^{\eta}$ as $\mathbf{x}_k = (\tilde{\mathbf{p}}_k, v_k) \in \mathbb{R}^4$ for $k = 1, 2, \ldots, m$. 

Our framework is agnostic to the specific anomaly detection algorithm employed, requiring only that the detector produces normalized anomaly scores for each feature. The detector $\hat{f}$ evaluates each new observation $\mathbf{x}_{m+1}$ from the data stream and produces a vector $\boldsymbol{a}_{m+1} \in [0,1]^4$:
\begin{equation}
    \boldsymbol{a}_{m+1} = \hat{f}(\mathbf{x}_{m+1}),
\end{equation}
where each element of the vector represents the anomaly score for the corresponding feature, with values closer to 1 indicating greater deviation from normal behavior. Subsequently, a binary anomaly label vector $\boldsymbol{y}_{m+1} \in \{0,1\}^4$ is derived from the score vector:
\begin{equation}
   \boldsymbol{y}_{m+1}^j = \mathbf{1}[\boldsymbol{a}_{m+1}^j > \theta^j]
\end{equation}
where $\theta^j$ is a threshold parameter for feature $j$ and $\mathbf{1}[\cdot]$ is the indicator function. Each component $\boldsymbol{y}_{m+1}^j = 1$ indicates that feature $j$ is anomalous in the new observation. Both $\boldsymbol{a}_{m+1}$ and $\boldsymbol{y}_{m+1}$ serve as inputs to the explainer component.

\subsubsection{The Explainer Component}
The semantic meaning of our proposed features enables the interpretation of anomaly detector outputs through failure localization and incident severity classification, facilitating business impact assessment.

\paragraph{Failure Localization.} 
The explainer leverages feature mapping to distributed system components to identify failure origins. It takes the binary anomaly label vector $\boldsymbol{y}_{m+1}$ and a set of predefined localization rules $\mathcal{L}$ to determine where in the system architecture the failure likely originated:
\begin{equation}
    \hat{g}_{\mathcal{L}}(\boldsymbol{y}_{m+1}, \mathcal{L}) \rightarrow \text{failure location}.
\end{equation}

For instance, the most basic localization categorizes failures as internal or external to the CSM infrastructure. The feature space directly facilitates this localization because different features naturally map to different system components.

\paragraph{Incident Severity Classification.} 
The explainer also provides incident severity classification by analyzing anomaly score combinations across features for detected anomalies. The classification function is represented as:
\begin{equation}
    \hat{g}_{\mathcal{C}} ( \boldsymbol{a}_{m+1}, \boldsymbol{y}_{m+1}, \mathcal{C}) \rightarrow \text{incident severity},
\end{equation}
where $\mathcal{C}$ represents domain knowledge about the relationship between feature deviation patterns and incident severity classification.

\paragraph{Business Impact Assessment.}
The combination of incident severity classification and failure localization enables business impact assessment. In distributed payment systems, business impact varies significantly even for incidents of similar severity, depending on contextual factors such as timing (e.g., during peak transaction hours) and duration (e.g., brief spikes vs. sustained outages). The framework provides both severity and localization information, allowing operational teams to assess the magnitude of business consequences and determine appropriate response strategies. For instance, an incident classified as major severity with internal localization requires direct intervention, while the same severity incident with external localization necessitates coordination with external service providers. This differentiation allows teams to prioritize responses based on business impact rather than technical metrics alone.

\section{Failure Detection on TARGET Instant Payment Settlement} \label{sec:experiments}
We evaluate our anomaly detection framework on the TARGET Instant Payment Settlement (TIPS) system~\citep{renzetti2021tips,arcese2021real,caricato2022tips}, addressing three research questions: whether our approach can (1) detect system failures, (2) localize failures in distributed system components, and (3) classify incident severity correctly.

Our evaluation uses both production data from a Network Service Provider incident and controlled scenarios with systematic anomaly injection in a TIPS testing environment. Since we treat the anomaly detector as a black box, we focus on whether known disturbances produce meaningful variations in our engineered features.

\subsection{Real-Time Feature Extraction Architecture} \label{sec:feature_pipeline}
\begin{figure}[h!]
    \centering
    \includegraphics[width=0.95\linewidth]{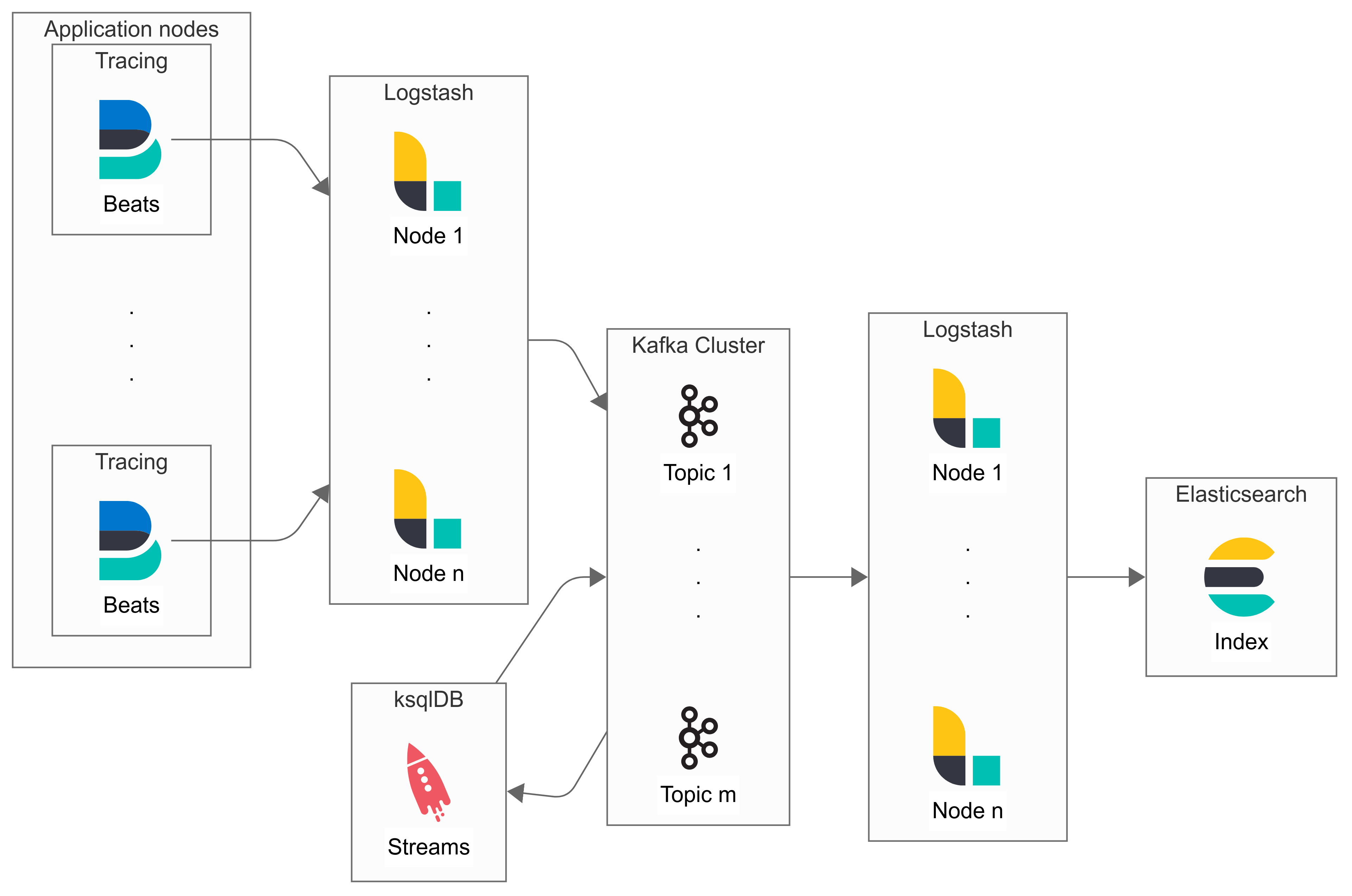}
    \caption{\small Real-time feature extraction pipeline architecture. Application nodes generate message traces, which are collected by Beats agents and processed by Logstash nodes. Messages flow through Kafka for buffering, where ksqlDB correlates messages to compute processing times. The processed features are then indexed in Elasticsearch for both storage and real-time analysis.}
    \label{fig:pipeline_architecture}
\end{figure}

Real-time computation of the proposed features requires dedicated infrastructure to process payment message traces as they occur. As illustrated in Figure~\ref{fig:pipeline_architecture}, we implemented an event-driven architecture comprising multiple distributed components for data collection, transformation, aggregation, and storage.

The pipeline begins at application nodes within the TIPS infrastructure, where ISO 20022 payment message traces are captured by Elastic Beats agents and forwarded to a Logstash cluster for initial preprocessing. An Apache Kafka cluster provides message buffering to handle variable transaction volumes without data loss. The core feature computation occurs in ksqlDB, which performs real-time stream processing to calculate processing times ($\delta_1$, $\delta_2$, $\delta_3$) by matching corresponding messages and computing their time differences.

The processed features are indexed in Elasticsearch in two forms: raw data for historical analysis and model training, and aggregated metrics (with frequency $\eta$) for real-time anomaly detection. The aggregation process incorporates the transaction volumes feature ($v$), completing the multivariate time series representation of system behavior. This distributed architecture supports horizontal scaling of each component based on workload requirements, ensuring the feature extraction process maintains performance even during peak transaction periods.

\subsection{Experimental Testbed Design} \label{sec:simulation}
\begin{figure}[h!]
   \centering
   \includegraphics[width=0.98\linewidth]{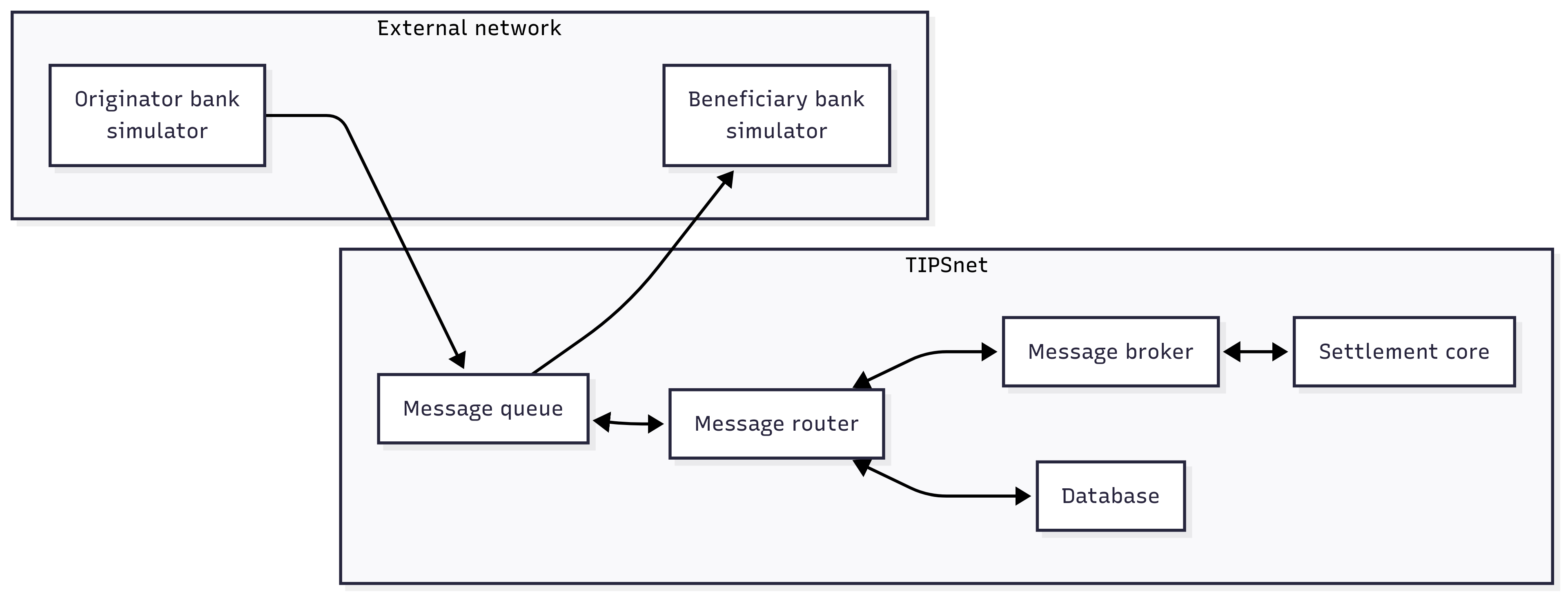}
   \caption{\small Architecture of the instant payment experimental testbed. External simulators representing the originator and beneficiary banks connect to the TIPSnet testing environment. Payment messages flow through the message queue and router to the core components. Anomalies can be injected at various points in the processing flow.}
   \label{fig:testbed_architecture}
\end{figure}

To systematically evaluate our anomaly detection framework across different scenarios, we developed an experimental testbed that generates realistic instant payment traffic with controlled anomaly injection capabilities. As shown in Figure~\ref{fig:testbed_architecture}, the testbed consists of two external simulators representing originator and beneficiary banks connected to a test instance of the TIPS platform that mirrors the production environment.

These simulators, positioned in a network external to TIPSnet to realistically model Network Service Provider latency, exchange payment messages that follow processing flows identical to those of real transactions within the TIPS platform. The simulator is configured with message probability distributions matching production traffic patterns, implementing a multi-threaded application that emulates multiple originators continuously generating payment streams at realistic volumes.

During test execution, we inject controlled disturbances into key TIPS components shown in Figure~\ref{fig:testbed_architecture}, including the message router, message broker, and settlement core. This testbed enables simulation of various anomaly types, from processing slowdowns and network issues to simultaneous multi-component failures.

To ensure experimental reproducibility and comprehensive result tracking, both traffic generation and anomaly injection are orchestrated through automated Ansible playbooks. The message traces from these simulations are processed through the feature extraction pipeline in Figure~\ref{fig:pipeline_architecture}, creating datasets stored in Elasticsearch indices.

\subsection{Validation on Real-World and Simulated Data} \label{sec:results}
We validate our framework using two datasets created with the feature extraction pipeline shown in Figure~\ref{fig:pipeline_architecture}: one containing production traces during a Network Service Provider incident, and the other containing controlled experiments from our testbed with systematic anomaly injection. We present the experimental configuration and evaluate detection performance across different incident scenarios.

\subsubsection{Datasets}
The first dataset consists of transaction traces collected from TIPS during August 2023, including a major Network Service Provider incident on August 11 that impacted most participants for approximately 15 minutes, causing processing slowdowns and payment timeouts. This production incident provides an authentic validation case under genuine operational conditions.

The second dataset is derived from controlled experiments conducted on our internal testbed, designed to reproduce anomalous scenarios that are either rare or difficult to isolate in production settings. We synthesized the following four stress scenarios:

\begin{enumerate}[label=\textbf{Scenario \arabic*.}, leftmargin=2.3cm, labelwidth=1.9cm, labelsep=0.1cm, align=left]
    \item \textbf{Mild Internal Stress} (7 min). Stress was applied to a core component of the TIPS system to evaluate sensitivity to sub-critical performance degradation.
    \item \textbf{Stress on Multiple Internal Components} (5 min). Several TIPS core components were stressed in parallel with moderate load levels.
    \item \textbf{External Participant Disturbances} (16 min). Disturbances impacting transactions from multiple participants outside the TIPS were simulated to assess the framework's ability to discriminate between internal and external failure origins.
    \item \textbf{Heavy Internal Stress} (21 min). Multiple TIPS core components were subjected to severe degradation, causing significant impact on transaction processing.
\end{enumerate}

\subsubsection{Anomaly Detection}
To implement the anomaly detector component, we utilize Elastic's anomaly detection capabilities~\cite{elk_anomaly_detection}, which provide a hybrid ensemble learning approach combining clustering, time series decomposition, Bayesian distribution modeling, and correlation analysis. The engine establishes a probabilistic baseline from historical data while dynamically adapting to new observations, computing deviation scores for each time bucket with multi-stage noise reduction and statistical significance ranking. All experiments were conducted in an unsupervised learning setting.

\paragraph{Anomaly Detector Configuration.}
We configure a multi-metric detection job analyzing all four features with frequency $\eta = 1$ second. The detector identifies significant drops in settled payment volume $v$ and unusually high mean values in processing times $\delta_1$, $\delta_2$, $\delta_3$.

Each anomaly detector generates anomaly scores on a 0-100 scale based on three factors: single bucket impact (probability against historical distribution), multi-bucket impact (trends across the previous 11 intervals), and anomaly characteristics (duration and intensity). A renormalization process recalibrates historical scores when significant anomalies occur. We normalize the final scores to $[0,1]$ and set anomaly thresholds of $0.25$ for volume $v$ and $0.4$ for processing times.

\paragraph{Rules for the explainer component.}
To instantiate the explainer component for experimental validation, we define two simple rulesets $\mathcal{L}$ and $\mathcal{C}$ based on TIPS architectural knowledge. The localization ruleset $\mathcal{L} = \{\text{L1, L2}\}$ distinguishes between internal and external failure origins by analyzing patterns in the binary anomaly label vector $\boldsymbol{y}$.

\begin{enumerate}[label=\textbf{Rule L\arabic*.}, leftmargin=2cm, labelwidth=1.4cm, labelsep=0.2cm, align=left]
    \item When anomalies occur in processing times $\delta_1$ or $\delta_3$ without corresponding anomalies in $\delta_2$, this pattern suggests that TIPS internal components are experiencing delays while external communication remains unaffected, indicating an internal failure origin.
    \item When anomalies appear exclusively in processing time $\delta_2$ while $\delta_1$ and $\delta_3$ remain normal, this indicates that TIPS processes messages normally but external entities exhibit delayed responses, suggesting an external failure origin.
\end{enumerate}

The incident classification ruleset $\mathcal{C}=\{\text{C1, C2, C3, C4}\}$ maps anomaly score patterns to incident severity levels.
\begin{enumerate}[label=\textbf{Rule C\arabic*.}, leftmargin=2cm, labelwidth=1.4cm, labelsep=0.2cm, align=left]
    \item Anomalies in processing times ($\delta_1$, $\delta_2$, or $\delta_3$) without corresponding volume drops indicate performance degradation. Critical thresholds (anomaly scores $\geq 0.75$) may cause payment timeouts.
    \item Anomalies in settled payment volume $v$ with anomaly score $0.25 \leq \boldsymbol{a}(v) < 0.5$ indicate minor incidents.
    \item Anomalies in settled payment volume $v$ with anomaly score $0.5 \leq \boldsymbol{a}(v) < 0.75$ indicate major incidents.
    \item Anomalies in settled payment volume $v$ with anomaly score $\boldsymbol{a}(v) \geq 0.75$ indicate critical incidents.
\end{enumerate}

\paragraph{Business Impact Evaluation.}
Business impact measures the effect of incidents on business operations and processes. We define business impact based on the number of transactions that fail to settle correctly due to processing times exceeding the critical threshold defined by the SCT Inst schema. For simplicity, we assume that the detector's anomaly scores already incorporate contextual factors such as timing and duration.

We establish four incident categories based on business impact severity. Performance degradation occurs when processing time increases remain below SCT Inst thresholds, resulting in negligible user impact as transactions complete within acceptable timeframes. Minor incidents affect a limited subset of transactions with timeout-related failures. Major incidents impact substantial transaction volumes, causing significant business disruption for most users. Critical incidents represent near-complete service outages where the majority of transactions fail to process successfully.

Table~\ref{tab:anomaly_conditions} shows representative examples of explainer component outputs when interpreting anomaly score vectors using the previously defined rules $\mathcal{L}$ and $\mathcal{C}$. For instance, the row with ID \verb|EIN| indicates a minor incident caused by extended processing times outside the TIPS system, impacting a limited subset of transactions and causing timeouts. The value $\{0\}$ for $\boldsymbol{a}(\delta_1)$ and $\boldsymbol{a}(\delta_3)$ indicates that no anomalies were detected in the binary anomaly label vector $\boldsymbol{y}$ for these features.

\begin{table*}[h!]
    \centering
    \small
    \begin{tabular}{>{\centering\arraybackslash}p{0.05\linewidth}|>{\centering\arraybackslash}p{0.1\linewidth}>{\centering\arraybackslash}p{0.1\linewidth}>{\centering\arraybackslash}p{0.1\linewidth}>{\centering\arraybackslash}p{0.1\linewidth}|>{\centering\arraybackslash}p{0.1\linewidth}>{\centering\arraybackslash}p{0.1\linewidth}>{\centering\arraybackslash}p{0.13\linewidth}}
        \toprule
        ID&  $\boldsymbol{a}(v)$ &$\boldsymbol{a}(\delta_1)$ & $\boldsymbol{a}(\delta_2)$ & $\boldsymbol{a}(\delta_3)$  &  Localization & Incident&Business impact\\
        \midrule
        \verb|EPC|&  \{0\}&\{0\}& $[0.75,1.0]$& \{0\}& External& Perf. degr.&No\\
        \verb|IPC|&  \{0\}&$[0.75,1.0]$& \{0\}& $[0.75,1.0]$& Internal& Perf. degr.&No\\ 
        \verb|EIN|&  $[0.25,0.5)$&\{0\}& $[0.75,1.0]$ & \{0\}& External  & Minor&Yes\\
        \verb|IIN|&  $[0.25,0.5)$&$[0.75,1.0]$& \{0\}& $[0.75,1.0]$& Internal  & Minor&Yes\\
        \verb|EIJ|&  $[0.5,0.75)$&\{0\}& $[0.75,1.0]$ & \{0\}& External  & Major&Yes\\ 
        \verb|IIJ|&  $[0.5,0.75)$&$[0.75,1.0]$& \{0\}& $[0.75,1.0]$& Internal  & Major&Yes\\
        \bottomrule 
    \end{tabular}
    \vspace{0.5em}
    \caption{\small Representative examples of anomaly score intervals and corresponding explainer outputs showing failure localization, incident severity, and business impact assessment. Score ranges indicate deviation magnitude from expected behavior, with higher values representing more significant anomalies.} \label{tab:anomaly_conditions}
\end{table*}

\subsubsection{Case Study: A Network Service Provider Incident}\label{sec:real_incident}
\begin{figure}
    \centering
    \begin{subfigure}[b]{0.49\linewidth}
        \centering
        \includegraphics[width=\linewidth]{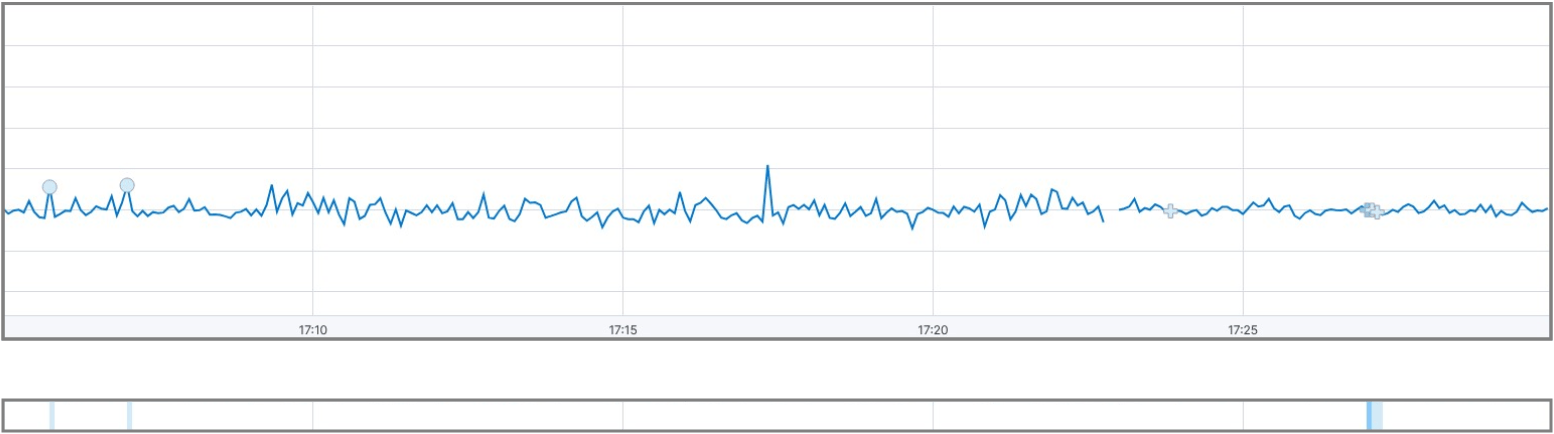}
        \caption{\small Processing time $\delta_1$}
        \label{fig:nsp_incident_scores_d1}
    \end{subfigure}
    \hfill
    \begin{subfigure}[b]{0.49\linewidth}
        \centering
        \includegraphics[width=\linewidth]{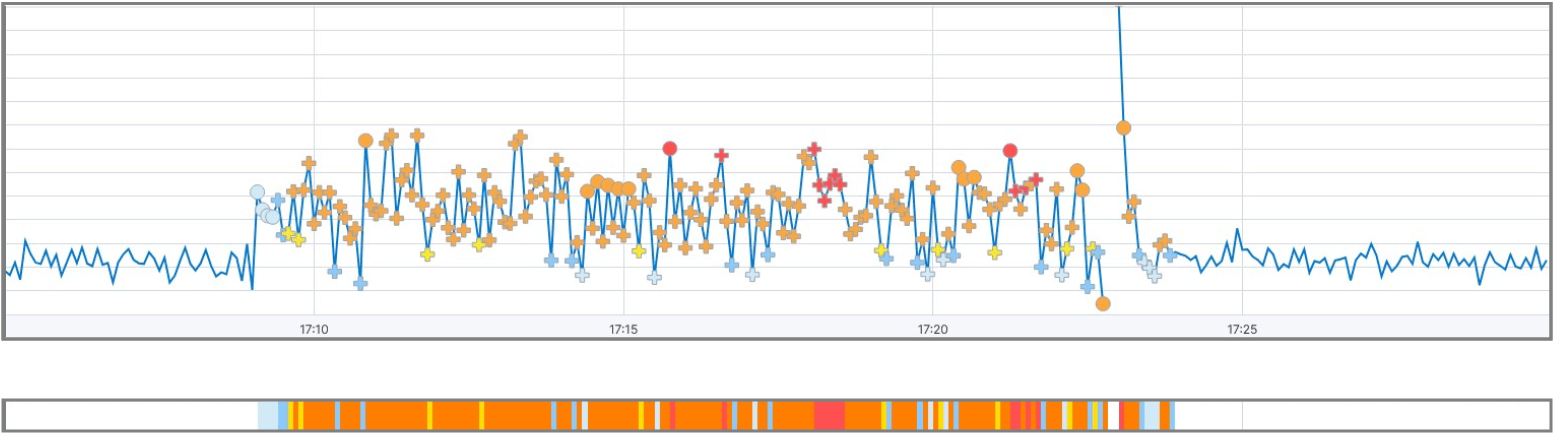}
        \caption{\small Processing time $\delta_2$}
        \label{fig:nsp_incident_scores_d2}
    \end{subfigure}
    
    \vspace{0.5cm}
    
    \begin{subfigure}[b]{0.49\linewidth}
        \centering
        \includegraphics[width=\linewidth]{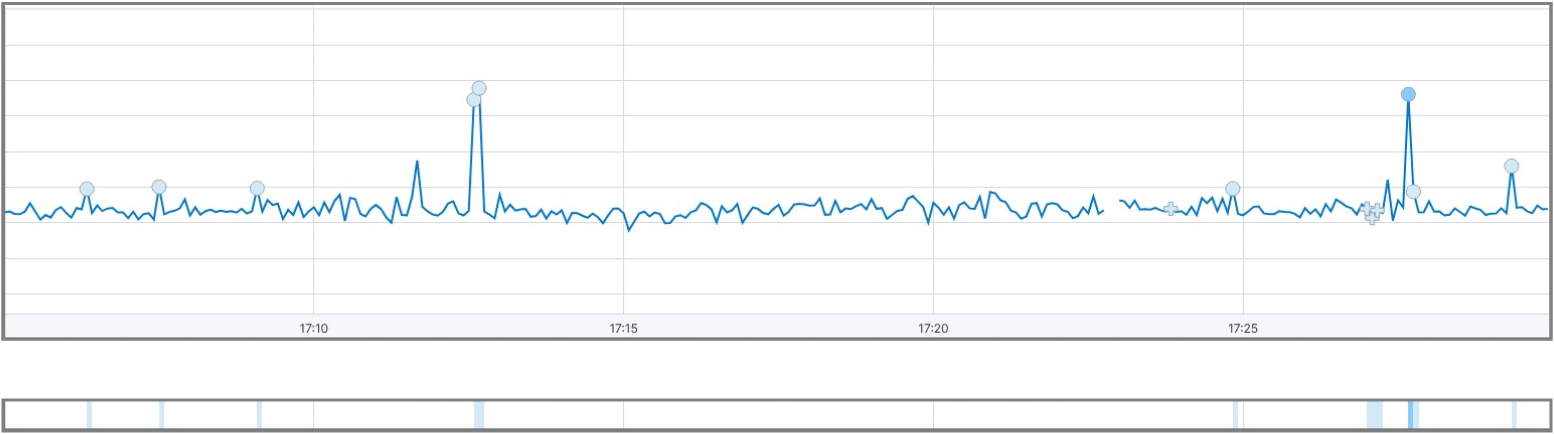}
        \caption{\small Processing time $\delta_3$}
        \label{fig:nsp_incident_scores_d3}
    \end{subfigure}
    \hfill
    \begin{subfigure}[b]{0.49\linewidth}
        \centering
        \includegraphics[width=\linewidth]{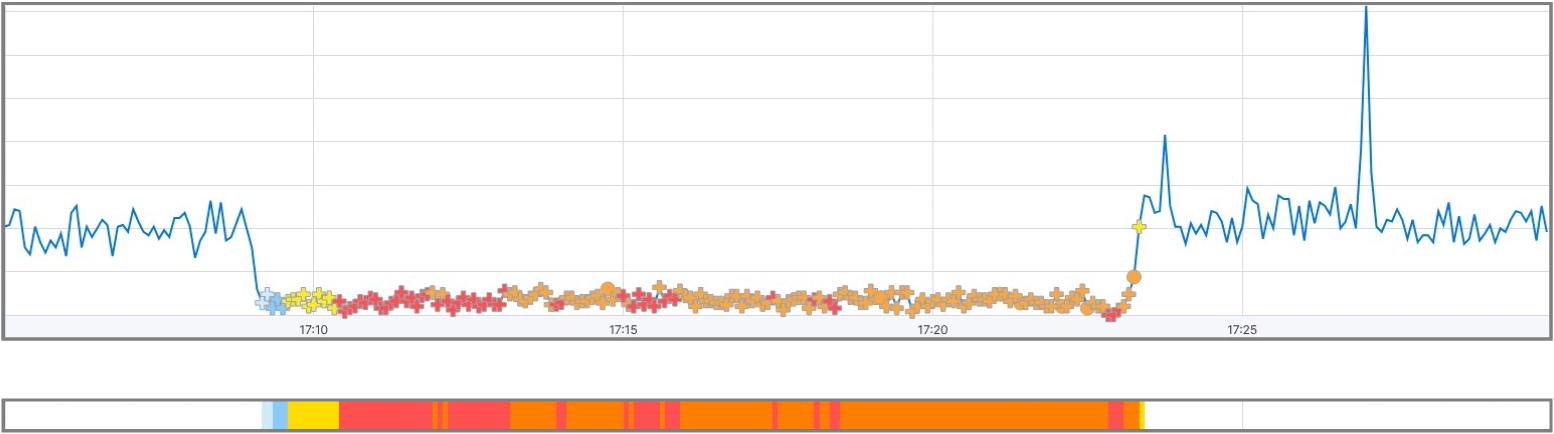}
        \caption{\small Settled instant payments $v$}
        \label{fig:nsp_incident_scores_v}
    \end{subfigure}
    \caption{\small Time series with corresponding anomaly scores for processing times $\delta_1$, $\delta_2$, $\delta_3$, and payment volume $v$ during the Network Service Provider incident. The plots illustrate changes in $\delta_2$ (external response time) and payment volume during the incident period, while internal processing times ($\delta_1$ and $\delta_3$) remain stable.} \label{fig:nsp_incident_features}
\end{figure}

The Network Service Provider (NSP) incident provides a validation of our framework's ability to distinguish between internal and external failure origins. Figure~\ref{fig:nsp_incident_features} illustrates the behavior of all four features during the NSP incident period. The time series data reveal that processing times $\delta_1$ and $\delta_3$ maintained relatively consistent values throughout the incident, with no significant anomalies detected. Conversely, the processing time $\delta_2$ exhibited notable increases, with values exceeding 10 seconds compared to typical sub-second response times under normal conditions. This coincided with a 78.62\% decrease in settled payment volume $v$.

The anomaly detector assigned elevated scores to both $\delta_2$ and volume $v$, while scores for $\delta_1$ and $\delta_3$ remained below detection thresholds. According to localization Rule L2, the anomaly pattern suggests an external failure origin based on the isolated anomaly in $\delta_2$. The severity classification under Rule C4 indicated a critical incident due to the substantial volume reduction.

Post-incident root cause analysis conducted by specialist teams confirmed that network connectivity issues at the NSP level disrupted communication between TIPS and multiple participants, which aligns with the indications provided by the framework.

\subsubsection{Evaluation of Controlled Anomaly Scenarios}
The controlled testbed experiments allowed for evaluation across different failure conditions that are difficult to reproduce in production environments. Table~\ref{tab:anomaly_scores} presents the maximum anomaly scores recorded for each feature during the four synthetic scenarios. Each scenario exhibited different patterns in the feature space.

\begin{table}[h!]
    \centering
    \begin{tabular}{>{\centering\arraybackslash}p{0.23\linewidth}>{\centering\arraybackslash}p{0.16\linewidth}>{\centering\arraybackslash}p{0.16\linewidth}>{\centering\arraybackslash}p{0.16\linewidth}>{\centering\arraybackslash}p{0.16\linewidth}}\toprule
        Scenario &$\max \boldsymbol{a}(v)$ & $\max \boldsymbol{a}(\delta_1)$ & $\max \boldsymbol{a}(\delta_2)$ & $\max \boldsymbol{a}(\delta_3)$  \\\midrule
        No. 1 &\cellcolor{normalbehavior} --- & \cellcolor{severitycritical}0.80 & \cellcolor{normalbehavior} --- & \cellcolor{severitycritical}0.85  \\
        No. 2 &\cellcolor{severitymoderate}0.43 & \cellcolor{severitycritical}0.82 & \cellcolor{normalbehavior} --- & \cellcolor{severitycritical}0.86  \\
        No. 3 &\cellcolor{severitymoderate}0.43 & \cellcolor{normalbehavior} --- & \cellcolor{severitycritical}0.95 & \cellcolor{normalbehavior} ---  \\
        No. 4 &\cellcolor{severityhigh}0.71 & \cellcolor{severitycritical}0.83 & \cellcolor{normalbehavior} --- & \cellcolor{severitycritical}0.87  \\
        \bottomrule
    \end{tabular}
    \vspace{0.5em}
    \caption{\small Maximum anomaly scores assigned by the anomaly detector to each feature during the time window in which each anomaly scenario was simulated. Cells are color-coded based on severity: scores above 0.75 indicate critical anomalies, 0.50-0.75 indicate major anomalies, and scores below 0.50 indicate minor anomalies. Entries marked with ``---'' indicate that no anomalies were detected for that feature in the corresponding scenario.} \label{tab:anomaly_scores}
\end{table}

Scenario 1 represented subcritical internal stress where processing delays occurred without affecting transaction settlement rates. The anomaly detector identified elevated values for $\delta_1$ and $\delta_3$ while volume remained stable. When applying Rule L1, this pattern suggested an internal origin, while Rule C1 categorized it as performance degradation with no immediate business impact. This scenario demonstrates the detection of early-stage performance issues before they affect service levels.

Scenario 2 exhibited internal processing delays in $\delta_1$ and $\delta_3$ that coincided with a minor volume reduction. Rule L1 indicated an internal origin, while Rule C2 categorized this as a minor incident due to the volume impact. This pattern may represent the evolution from performance degradation to measurable business impact.

Scenario 3 showed isolated $\delta_2$ anomalies accompanied by minor volume impact. The application of Rule L2 suggested an external origin, while Rule C2 categorized the incident severity similarly to the previous scenario. This pattern distinguishes external participant issues from internal system component failures.

Scenario 4 demonstrated severe internal degradation where $\delta_1$ and $\delta_3$ anomalies coincided with major volume reduction. The application of Rules L1 and C3 identified this as a major internal incident, impacting most users.

These experiments indicate that the engineered features based on processing times can reflect system state changes across different failure modes, potentially enabling assessment of failure localization and incident severity through pattern analysis.

\subsection{Discussion} \label{sec:discussion}
Experimental results provide compelling validation for the efficacy of our feature engineering approach for failure detection in instant payment infrastructures. We discuss the effectiveness of the anomaly detection framework, its comparative advantages over traditional monitoring approaches, practical implications for operations, current limitations, and future directions.

\subsubsection{Feature Engineering Effectiveness}
The real-world NSP incident and controlled scenario evaluations demonstrate the effectiveness of the proposed features in capturing the operational state of instant payment systems. These features successfully identify anomalies that correspond to observed disturbances across various failure modes.

The proposed approach significantly reduces mean time to response by providing immediate, contextually relevant insights about anomaly origins. Analysis of detected anomaly combinations enables the identification of incident origins as either internal or external to TIPS, while integration of domain knowledge allows for business impact assessment based on incident severity. This enables operators to prioritize responses according to both location and business impact, bypassing lengthy investigative processes that typically require specialized domain expertise and focusing remediation efforts directly on affected system components.

As demonstrated in the experimental phase, a critical anomaly does not always signify a critical incident with business impact. This framework enables reclassification of incident severity based on business impact with clear explanations that guide operator responses.

\subsubsection{Advantages Over Traditional Infrastructure Monitoring}
The resulting anomaly detection framework offers significant complementary capabilities to conventional infrastructure monitoring, functioning as an additional semantic layer bridging technical metrics with business process visibility rather than as a replacement. Traditional monitoring excels at resource utilization tracking and assessment of system component health and performance, but faces challenges with incidents involving external elements.

The NSP incident exemplifies this limitation: such incidents manifest as business-level anomalies (reduced settled payments) without corresponding infrastructure alerts, where component-level monitoring shows normal operation while integrated service experiences significant degradation. The proposed anomaly detection framework addresses this observability gap by leveraging engineering features that span component boundaries and capture end-to-end transaction flows, enabling detection and explanation of anomalies occurring outside directly monitored infrastructure.

Our approach detects incidents in distributed systems while providing actionable insights that guide remediation efforts through failure localization and severity classification, enabling incident characterization for root cause analysis, rapid routing to appropriate IT teams, and immediate business impact assessment.

\subsubsection{Implementation Considerations and Real-Time Operations}
The anomaly detection framework is agnostic to the specific anomaly detection algorithm used. However, to support real-time operations, both anomaly detector and explainer must operate within strict time constraints, producing results in a cumulative time $\varepsilon$ less than the sampling frequency ($\varepsilon < \eta$), thus requiring algorithms capable of supporting real-time anomaly detection.

The explainer component fundamentally relies on the semantic richness of the proposed features. Since these features inherently map to different process phases executed on different system components, they naturally support both localization and incident severity classification through an explainability by design approach, contrasting with post-hoc explanation methods that attempt to explain opaque model decisions after detection.

Unlike AIOps solutions that process thousands of real-time metrics, the proposed feature engineering is computationally lightweight, focusing on compact features that represent business-relevant system states. Feature engineering at the CSM level enables real-time payment state tracking across distributed participants, despite the inherently fragmented nature of global transaction state where each participant observes only their relevant ISO 20022 messages. 

Beyond feature extraction, the raw transactions reconstructed by the pipeline (Figure~\ref{fig:pipeline_architecture}) directly accelerate operators' daily investigations, reducing customer support response times. Operators gain immediate access to all messages composing a specific payment, eliminating laborious manual correlation.

\subsubsection{Limitations and Future Directions}
Our current approach cannot anticipate failures that occur without preceding warning signs, such as complete network disconnections that immediately terminate all connections. Such catastrophic events provide no gradual performance degradation for advance detection, highlighting the need for complementary monitoring mechanisms.

The level of detail presented in this study does not directly identify faults but, by allowing failure localization, reduces the time required to initiate root cause analysis (RCA), which aims to map an incident to its underlying fault. Future research could explore combining insights from the proposed approach with traditional infrastructure monitoring data to facilitate comprehensive RCA. By correlating business-level anomalies with low-level system metrics, this integration may enable automated remediation strategies where the proposed framework establishes a semantic layer that contextualizes anomalies within business process flows and triggers appropriate corrective actions.

While our implementation focused on the CSM perspective, the underlying anomaly detection framework demonstrates considerable flexibility and may be adapted for other SCT Inst process actors, such as individual Payment Service Providers, creating opportunities for monitoring at different ecosystem levels.

Another promising extension involves integrating our approach with Large Language Models (LLMs) to enhance operational usability through natural language interfaces, providing operators with enriched explanations that incorporate additional contextual information and recommended corrective actions in natural language, thereby reducing the technical expertise required for effective incident response.

\section{Conclusion} \label{sec:conclusion}
This research introduces a novel feature engineering approach for anomaly detection in instant payment infrastructures, extracting meaningful insights from ISO 20022 message exchanges to create compact representations of payment system state. By calculating processing times between consecutive messages, we achieve observability of the SCT Inst settlement process in a distributed context where no single actor has complete transaction visibility.

Our approach focuses on business-relevant system states, enabling failure detection beyond directly monitored components. We treat failure detection as an anomaly detection problem, leveraging the assumption that failures generate irregular behaviors by increasing processing times or reducing settled payment volumes. The interpretation of detected anomalies enables early failure detection and localization, allowing incident classification that directly informs business impact assessment.

The framework provides algorithm-agnostic anomaly detection, where engineered features inherently support reasoning about system behavior without requiring post-hoc interpretation. This enables incident characterization that reduces reliance on specialized domain expertise for initial diagnosis, thereby reducing mean time to response by facilitating routing to appropriate IT teams and guiding remediation efforts.

Experimental evaluation on the TARGET Instant Payment Settlement (TIPS) system demonstrates effectiveness across both real-world incidents and controlled simulations. While implemented for a Clearing and Settlement Mechanism following the SEPA Instant Credit Transfer scheme, the core approach generalizes to other distributed financial systems by providing a semantic layer that enhances system behavior awareness and enables more effective AI-powered monitoring with increased trust in detection results.

\section*{Acknowledgments}
The opinions expressed in this paper are personal and should not be attributed to the Bank of Italy.

The author would like to express gratitude to all developers and technical administrators who provided valuable support during the experimental setup phase within their respective areas of expertise.

\bibliographystyle{unsrtnat}
\bibliography{references}

\end{document}